\documentclass[a4paper,fleqn]{cas-dc}

\usepackage[numbers]{natbib}
\usepackage{seqsplit}
\usepackage{hyperref}
\usepackage{graphicx}

\def\tsc#1{\csdef{#1}{\textsc{\lowercase{#1}}\xspace}}
\tsc{WGM}
\tsc{QE}
\tsc{EP}
\tsc{PMS}
\tsc{BEC}
\tsc{DE}

\begin{document}
\let\WriteBookmarks\relax
\def\floatpagepagefraction{1}
\def\textpagefraction{.001}
\shorttitle{Vibe Medicine}
\shortauthors{Z. Wu et~al.}

\title [mode = title]{Vibe Medicine: Redefining Biomedical Research Through Human-AI Co-Work}                      

\author[1]{Zihao Wu}\cormark[1]
\author[1]{Steven Xu}\cormark[1]
\author[2]{Bowen Chen}\cormark[1]
\author[2]{Shaowen Wan}\cormark[1]
\author[1]{Yiwei Li}
\author[1]{Wei Ruan}
\author[3]{Yanjun Lyu}
\author[1]{Siyuan Li}
\author[3]{Dajiang Zhu}
\author[1]{Tianming Liu}
\author[2]{Lin Zhao}[orcid=0000-0003-2567-627X]\cormark[2]
\ead{lin.zhao.1@njit.edu}

\affiliation[1]{School of Computing, University of Georgia, Athens, GA 30602, USA}
\affiliation[2]{Department of Biomedical Engineering, New Jersey Institute of Technology, Newark, NJ 07102, USA}
\affiliation[3]{Department of Computer Science and Engineering,  University of Texas at Arlington, Arlington, TX 76019, USA}

\cortext[cor1]{Equal contribution}
\cortext[cor2]{Corresponding author}

\begin{abstract}
With the emergence of large language models (LLMs) and AI agent frameworks, the human-AI co-work paradigm known as Vibe Coding is changing how people code, making it more accessible and productive. In scientific research, where workflows are more complex and the burden of specialized labor limits independent researchers and those in low-resource areas, the potential impact is even greater, particularly in biomedicine, which involves heterogeneous data modalities and multi-step analytical pipelines. In this paper, we introduce \textbf{Vibe Medicine}, a co-work paradigm in which clinicians and researchers direct skill-augmented AI agents through natural language to execute complex, multi-step biomedical workflows, while retaining the role of research director who specifies objectives, reviews intermediate results, and makes domain-informed decisions. The enabling infrastructure consists of three layers: capable LLMs, agent frameworks such as OpenClaw and Hermes Agent, and the OpenClaw medical skills collection, which includes more than 1,000 curated skills from multiple open-source repositories. We analyze the architecture and skill categories of this collection across ten biomedical domains, and present case studies covering rare disease diagnosis, drug repurposing, and clinical trial design that demonstrate end-to-end workflows in practice. We also identify the principal risks, such as hallucination, data privacy, and over-reliance, and outline directions toward more reliable, trustworthy, and clinically integrated agent-assisted research that advances research and technological equity and reduces health care resource disparities.
\end{abstract}


\maketitle
\section{Introduction}
\label{sec:intro}

\begin{figure*}[t]
    \centering
    \includegraphics[width=\textwidth]{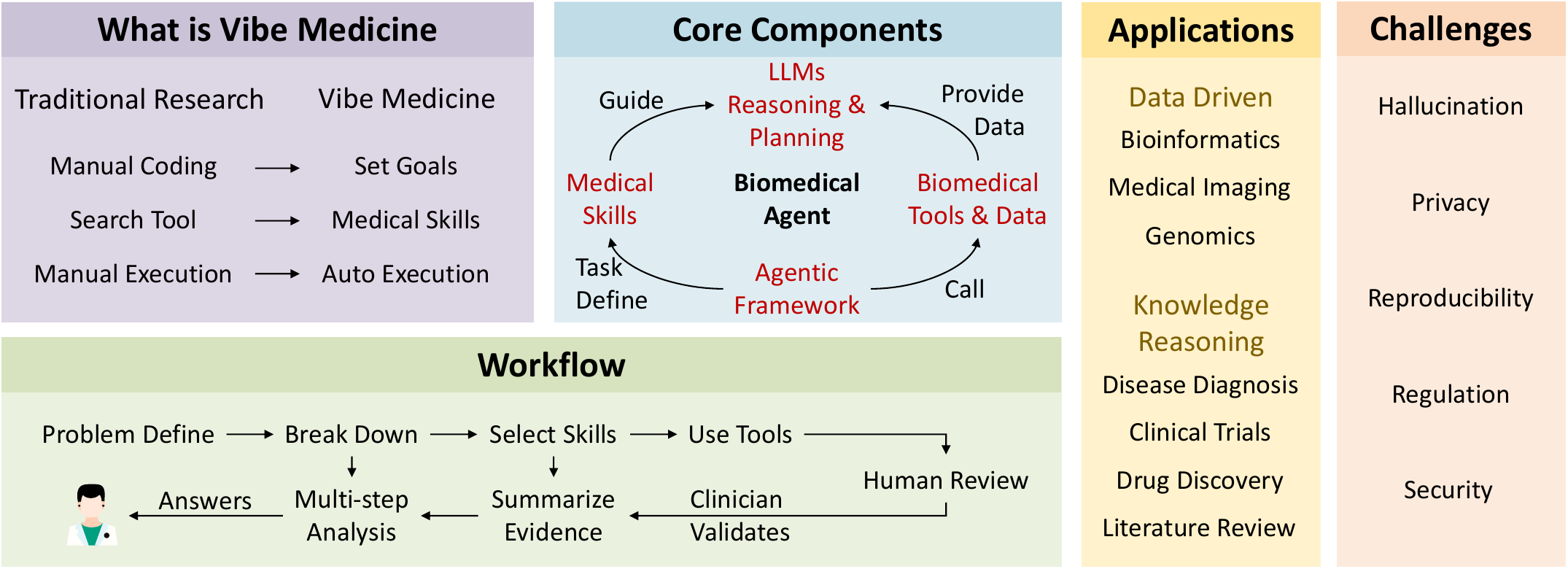}
    \caption{Conceptual overview of Vibe Medicine. The figure contrasts traditional research with Vibe Medicine and shows how LLMs, agent frameworks, medical skills, and biomedical tools and databases work together in a human-in-the-loop process, while also highlighting key applications and challenges.}
    \label{fig:workflow}
\end{figure*}

\begin{figure*}[t]
    \centering
    \includegraphics[width=1\textwidth]{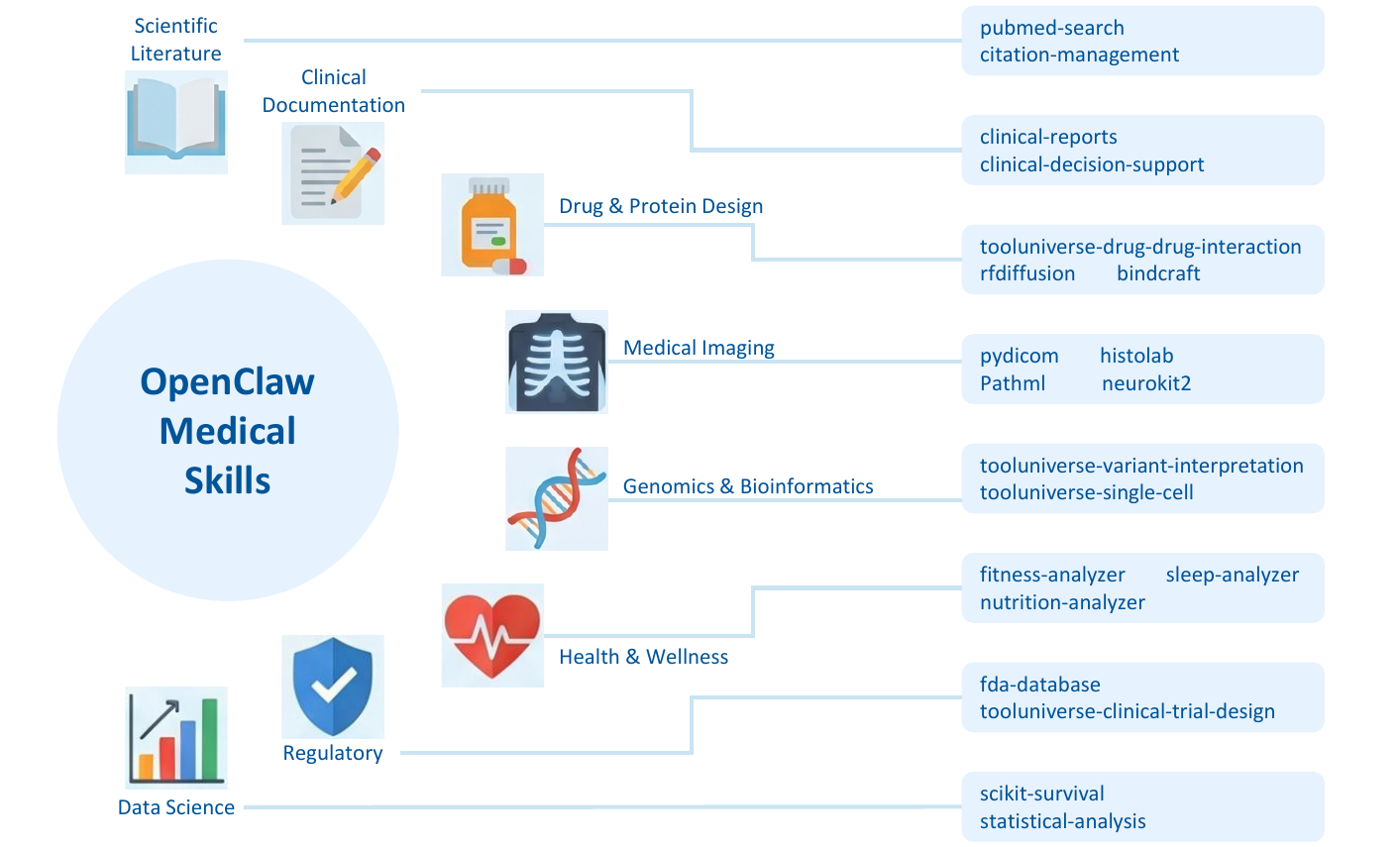}
    \caption{Overview of the OpenClaw medical skills collection, which spans the full stack of biomedical research workflows that Vibe Medicine is designed to support. The left panel organizes biomedical capabilities across major domains, covering a diverse range of biomedical data modalities and tasks. The right panel lists representative skills for each domain.}
    \label{fig:skills}
\end{figure*}

In February 2025, Andrej Karpathy introduced the term Vibe Coding\footnote{\url{https://x.com/karpathy/status/1886192184808149383}} to describe a human-AI co-work paradigm in software development: humans describe their goals with a high-level prompt in plain language and an AI model generates, refines, and debugs code through the conversation \cite{ge2025survey}. The vibe quickly spread widely into scientific research that frees researchers up to focus on the high-level big picture and idea while AI handles implementation details \cite{MooreTatonetti2025}. In biomedical research, where many researchers and clinicians have substantial domain expertise but limited formal programming training, this vibe is especially impactful in bridging the gap between a research idea and executable code \cite{LeeHuh2025}.
 
This paper introduces the concept of \textbf{Vibe Medicine}, which extends the human-AI co-work paradigm from code generation into full-scope biomedical research. As shown in Fig.~\ref{fig:workflow}, in Vibe Medicine, a clinician or researcher does not simply ask a model to write a script; instead, they describe a research objective, and a skill-augmented AI agent decomposes the objective into subtasks, selects and calls the appropriate skills and tools, executes multi-step analysis pipelines, and returns structured results. The human's role shifts from engineering programmer to research director focused on the big picture.
 
The infrastructure enabling this paradigm has emerged rapidly. On the model side, LLMs such as Claude and GPT now support extended reasoning, tool use, and code execution within a single session. On the framework side, open-source agent platforms such as OpenClaw and Hermes provide persistent, locally hosted agents that can autonomously plan, execute, and revise multi-step workflows through chat interfaces \cite{OpenClawGitHub,HermesAgent2026}. On the capability side, OpenClaw medical skills collection \cite{OpenClawMedicalSkills,aipoch_medical_skills_repo} includes more than 1,000 curated, domain-specific skills from multiple open-source repositories. Fig.~\ref{fig:skills} summarizes these skills across biomedical domains, covering scientific literature, clinical documentation, drug and protein design, medical imaging, genomics and bioinformatics, health and wellness, regulatory compliance, data science, and more. Each skill is packaged as a self-contained \texttt{SKILL.md} specification that teaches the agent what to do, which tools or APIs to call, and how to structure its output. Together, these three layers create the bedrock for Vibe Medicine to become practically feasible.
 
This development of Vibe Medicine redefines biomedical research and makes it more accessible. Modern radiology research increasingly extends beyond image interpretation into radiomics, radiogenomics, multi-omics data fusion, clinical trial design, and translational analysis, all of which require programming skills, database access, statistical expertise, and cross-disciplinary knowledge that few individual researchers possess \cite{Faghanietal2025,Salehietal2025}. Medical-skill-augmented AI agents can lower these barriers, especially for independent researchers and institutions in low-resource areas.
 
This paper makes four contributions. In Sections~\ref{sec:intro} and~\ref{sec:Background}, we define Vibe Medicine as a human-AI co-work paradigm and distinguish it from both Vibe Coding and conventional agent-based automation. In Section~\ref{sec:med-skill}, we analyze the architecture and skill categories of the OpenClaw medical skills collection, describing how more than 1,000 skills are organized, composed, and deployed. In Sections~\ref{sec:skill-cap} and~\ref{sec:case}, we survey the skill capabilities across ten biomedical domains and present concrete case studies that demonstrate end-to-end Vibe Medicine workflows in practice. In Sections~\ref{sec:challenges} and~\ref{sec:future}, we identify the principal challenges and outline future directions toward more reliable, trustworthy, and clinically integrated agent-assisted biomedical research.

\section{Background \& Related Work}
\label{sec:Background}

\subsection{Vibe Coding and Agentic Engineering}
Recent changes in AI-assisted work can be understood through the contrast between Vibe Coding and agentic engineering. At its core, Vibe Coding refers to a mode of collaboration in which a human specifies the goal, task, or desired outcome in natural language, and the model takes over much of the implementation, including code generation, revision, debugging, and completion. The emphasis is on high-level intent specification: the human states what should be done, and the model tries to make it happen.

By contrast, agentic engineering is concerned not only with producing an output, but with organizing the model inside a task-execution loop. In this setting, the model must decompose the task, decide what to do next, call external tools, inspect intermediate results, and revise its strategy when needed. The focus therefore shifts from making the model generate plausible content to making it act reliably within a structured process.

This distinction has direct implications for medicine. Many clinical and scientific tasks are not completed by generating fluent text alone; they require evidence retrieval, database access, integration of heterogeneous information, and explicit space for human review and correction. In other words, medical work often requires a system not only to ``say'' something, but also to search, execute, and verify. For this reason, Vibe Medicine is better situated in the context of agentic engineering: its central concern is not whether a model can produce convincing medical language, but whether it can complete traceable, inspectable, and correctable tasks within real biomedical workflows \cite{moore2025prompt}.

\subsection{AI Agent Frameworks}
AI agent frameworks emerged from the need to move beyond fixed execution chains. Conventional pipelines are usually specified in advance, with predetermined inputs, outputs, and step order. Workflows may allow branching and conditional logic, but they still follow a largely predefined structure. An agent differs in that it can decide which action to take next on the basis of intermediate observations, tool outputs, or user feedback. In current LLM-based systems, this adaptive loop is commonly described through four recurring functions: planning, action, memory, and reflection \cite{Xietal2023,XuSankar2025}. Planning selects the next subgoal, action executes external operations, memory preserves relevant context across steps, and reflection supports error detection and revision. Reasoning-and-acting frameworks such as ReAct made this interaction pattern explicit and have strongly influenced subsequent agent designs \cite{Yaoetal2023}.

In implementation terms, current agent frameworks can be grouped into three broad forms. The first emphasizes single-agent tool use and execution control, with representative examples including general-purpose execution agents such as AutoGPT and coding or research assistants such as Codex, Claude Code, and Cursor. The second emphasizes multi-agent collaboration through role specialization, task decomposition, and communication, with representative examples such as CAMEL and MetaGPT \cite{Lietal2023CAMEL,Hongetal2024MetaGPT}. The third emphasizes reusable capability modules that organize external tools, domain knowledge, and standard procedures, with representative examples such as HuggingGPT, where a LLM acts as a controller over a collection of expert models \cite{Shenetal2023HuggingGPT}.

\subsection{Medical AI Agents}
Medical AI agents extend these general architectures into clinical and biomedical settings by linking language-based reasoning with domain-specific action. Rather than only answering medical questions, they are designed to execute meaningful subtasks such as evidence retrieval, patient-to-trial matching, risk calculation, variant interpretation, report drafting, or multimodal case analysis. Recent reviews organize this growing literature into several broad categories, including patient-facing agents, workflow automation assistants, decision-support agents, and research-oriented biomedical agents \cite{Njeietal2026,XuSankar2025}. Although these categories overlap in practice, they capture an important shift: medical AI is moving from isolated prediction models toward systems that coordinate multiple subtasks, multiple tools, and multiple evidence sources.

Representative systems illustrate this transition. Patient-facing and communication-oriented agents emphasize triage, education, and longitudinal interaction. Workflow assistants target documentation, coding, scheduling, and structured information extraction. Decision-support agents increasingly integrate heterogeneous inputs such as text, structured records, images, and physiological data to support diagnosis, prognosis, and treatment planning. Research-oriented agents operate over biomedical databases, literature collections, and code execution environments to support genomics, drug discovery, and translational analysis \cite{Njeietal2026,XuSankar2025}. AgentMD provides a concrete example of tool-grounded medical agency by coupling an LLM with a large library of executable clinical calculators and demonstrating improved clinical risk prediction relative to prompting alone \cite{Jinetal2025AgentMD}. In radiology, recent perspective articles suggest that agentic systems may be particularly relevant for report generation, workflow orchestration, multimodal reasoning, and the integration of imaging findings with longitudinal clinical context, while also emphasizing unresolved issues in evaluation, hallucination control, privacy, and deployment \cite{Faghanietal2025,Salehietal2025,Tzanisetal2026,Dietrich2025}.

\subsection{LLMs in Medicine}
The emergence of medical LLMs laid the foundation for this line of work. Models such as Med-PaLM, BioGPT, PMC-LLaMA, HuatuoGPT, and related biomedical variants demonstrated that domain adaptation can improve medical question answering, scientific text understanding, and instruction following in healthcare settings \cite{Singhaletal2023,Luoetal2022,Wuetal2024,Zhangetal2023HuatuoGPT}. These models can themselves be grouped into several overlapping directions: general-domain LLMs adapted to medicine through instruction tuning, biomedical language models pretrained on domain corpora, clinically aligned chat models, and open-source medical assistants designed for downstream customization. Collectively, they established that language models can encode substantial amounts of biomedical knowledge and can interact more naturally with clinicians and researchers than earlier task-specific models.

However, most early medical LLM studies primarily evaluated prompt-based interactions over static model parameters, sometimes augmented with retrieval but often without reliable execution over external tools. This creates an important gap between knowing medicine and doing medical work. Many real tasks require current literature, structured database queries, image or omics pipelines, guideline-aware synthesis, and explicit provenance. A model may know the general treatment paradigm for a disease, yet still fail if the actual task requires querying active trials, retrieving an FDA label, calculating a validated risk score, or combining imaging, pathology, and genomics findings in a single report. As a result, even strong medical LLMs remain limited when asked to support end-to-end workflows that depend on up-to-date external evidence, quantitative analysis, or reproducible action sequences. This limitation is one of the main reasons the field has shifted from medical LLMs to medical agents.

\subsection{Tool-Augmented LLMs for Science}
Tool augmentation became a natural response to the limitations of prompt-only medical LLMs. Once tasks require up-to-date literature, structured databases, numerical calculation, or executable analysis, language modeling alone is no longer sufficient. Tool-augmented systems address this problem by allowing models to call external resources such as search engines, APIs, scientific software, and code execution environments. In biomedicine, this shift is particularly important because many high-value tasks depend on real-time access to PubMed, ClinicalTrials.gov, FDA labels, pathway databases, omics pipelines, or laboratory analysis tools.

Recent systems illustrate several complementary directions. ToolUniverse emphasizes unified connectivity across a large collection of tools, datasets, and analysis resources \cite{Gaoetal2025ToolUniverse}. Biomni advances a more integrated biomedical agent paradigm that combines reasoning, code execution, and a biomedical knowledge layer \cite{Huangetal2025Biomni}. HuggingGPT demonstrates another influential design in which a LLM acts as a controller that selects and invokes expert models to solve complex tasks \cite{Shenetal2023HuggingGPT}. In a domain-specific direction, ChemCrow shows how tool augmentation can extend LLMs from general language processing to more operational scientific tasks in chemistry \cite{Branetal2024ChemCrow}. This perspective is useful for Vibe Medicine because it highlights that medically meaningful agent behavior does not arise from the language model alone, but from the interaction among reasoning, tools, workflows, data sources, and human supervision.

\section{OpenClaw Medical Skills: Architecture \& Design}
\label{sec:med-skill}

\subsection{Skill Design Philosophy}

OpenClaw adopts a multi-skill design in which each skill is defined as a self-contained functional unit that encapsulates a specific capability. In practice, a skill typically bundles together three essential elements: domain knowledge and workflows, access to external tools or APIs, and structured output conventions. This design allows the agent to treat complex medical or scientific tasks not as a single monolithic problem, but as a collection of smaller, well-defined operations that can be invoked when needed. As a result, skills are independently reusable, easier to inspect and maintain, and more straightforward to integrate into larger workflows~\cite{toolformer, gorilla}.

The atomic representation of a skill is a standardized specification file, typically \texttt{SKILL.md}. This file serves both as a human-readable description of what the skill does and as an executable interface that tells the agent how and when to use it. In other words, \texttt{SKILL.md} provides a common language between developers, domain experts, and AI agents. This abstraction is important because it lowers the barrier to extending the system: instead of retraining a large model whenever a new biomedical function is needed, developers can add or update a skill in a modular manner~\cite{langchain, openai_functions}. For readers unfamiliar with AI systems, this can be understood as packaging expert knowledge and tool usage into portable functional units, allowing the broader system to scale while remaining interpretable and maintainable.

\subsection{Skill Categories}

The OpenClaw medical skills collection includes more than 1,000 specialized skills spanning more than a dozen biomedical domains. These domains include general academic research tools, clinical and medical utilities, scientific databases, bioinformatics, omics and computational biology, ClawBio pipelines, the BioOS extended suite, and data science. This structure is necessary because biomedical work is inherently heterogeneous: answering a clinical question, analyzing sequencing data, searching the literature, and preparing a regulatory summary each require different knowledge sources and computational procedures.

This structure is not merely a directory of tools. Rather, it provides a structured view of the system's capability space, enabling the agent to identify which type of skill is appropriate for a given user request. For instance, a clinician-oriented question may trigger literature, clinical reasoning, and database-related skills, whereas a translational research task may require combining omics analysis, bioinformatics pipelines, and statistical utilities. By organizing skills into explicit categories, OpenClaw makes its biomedical scope more understandable to users and more manageable for the agent, while also supporting both highly specialized tasks and broader cross-domain workflows~\cite{react, autogpt, OpenClawMedicalSkills, aipoch_medical_skills_repo}.

\subsection{Skill Composition and Orchestration} 

A key strength of the framework is that individual skills are not used in isolation, but can be composed into multi-step pipelines that mirror real biomedical reasoning and research processes. Many medical tasks are naturally sequential: a user may first need to retrieve prior studies, then extract or annotate relevant biological entities, then interpret the results in a clinical or scientific context, and finally summarize the findings into a report. OpenClaw supports this type of workflow by allowing one skill's output to become the input to the next, thereby transforming a collection of isolated tools into a coordinated problem-solving system.

For example, a task may begin with literature retrieval, proceed to variant or biomarker annotation, and end with clinical interpretation or report generation. More broadly, skills can be chained across different domains, such as combining database querying, bioinformatics analysis, evidence summarization, and writing assistance within a single pipeline. This composition mechanism is important because it moves the system beyond single-function execution toward end-to-end task solving~\cite{react, voyager}.

\subsection{Installation and Deployment}

The deployment design emphasizes low-friction adoption so that users can access biomedical capabilities without substantial engineering effort. Skills can be installed and managed through a unified ClawHub registry, loaded either globally or within a local workspace, and integrated into lightweight runtime environments such as NanoClaw containerized agent systems. This deployment model is especially important in biomedical settings, where end users may include researchers, students, or clinicians who are not necessarily experienced in software engineering or AI infrastructure.

From a practical perspective, this design lowers the barrier to entry in two ways. First, it simplifies access to specialized capabilities by allowing users to install only the skills relevant to their tasks, rather than deploying an entire complex platform from scratch. Second, containerized environments such as NanoClaw improve portability and reproducibility, making it easier to run the same skill configuration across different machines or projects. Together, these choices make OpenClaw more accessible, easier to extend, and better suited for real-world biomedical use, where usability and deployment simplicity are often as important as raw model capability~\cite{docker, kubernetes, OpenClawMedicalSkills}.

\section{Skill Capabilities in Medicine}
\label{sec:skill-cap}

\subsection{Scientific Literature \& Reference Management}
\label{sec:literature}

Scientific literature search and reference management form the entry point for most biomedical research workflows, and the OpenClaw medical skills collection addresses this domain through a layered set of capabilities spanning multi-source search, evidence synthesis, citation management, systematic review support, and trend analysis. Representative skills include \texttt{pubmed-search}, \texttt{arxiv-search} and \texttt{biorxiv-database} for preprint retrieval, \texttt{\seqsplit{literature-review}} for multi-database reviews with deduplication and evidence grading, \texttt{citation-management} for reference organization and format conversion, and \texttt{\seqsplit{systematic-review-screener}} for title-abstract screening against inclusion/exclusion criteria. The AIPOCH medical-research-skills repository contributes complementary capabilities such as \texttt{\seqsplit{reference-search}}, \texttt{citation-network} for forward and backward citation tracing, \texttt{citation-chasing-mapping}, \texttt{journal-matchmaker}, and \texttt{semantic-scholar-database}~\cite{OpenClawMedicalSkills,aipoch_medical_skills_repo}.
 
\textbf{Multi-source literature search.} PubMed remains the primary index for biomedical literature, covering more than 37 million citations with Medical Subject Headings (MeSH) controlled vocabulary~\cite{pubmed2024}, but the growing importance of preprints in fast-moving fields such as genomics and infectious disease means that single-database search is often insufficient~\cite{fraser2021preprint}. The \texttt{biomedical-search} skill combines PubMed, preprint servers, clinical trial registries, and FDA drug labels into a single query, returning deduplicated results. A radiologist preparing a grant proposal on AI-assisted lung nodule detection can, in a single call, retrieve relevant peer-reviewed studies and machine-learning preprints that would otherwise require repeated searches across four separate interfaces.
 
\textbf{Evidence synthesis and review generation.} Systematic reviews and meta-analyses sit at the top of the evidence hierarchy in clinical medicine~\cite{murad2016evidence}, and the \texttt{literature-review} skill supports end-to-end review generation through multi-database search, key-finding extraction, thematic organization, and structured draft output with inline citations. The AIPOCH \texttt{literature-extensive-read} and \texttt{\seqsplit{literature-close-read}} skills decompose the process further into high-level corpus scanning and detailed per-paper critical analysis. A researcher drafting a paper on immunotherapy biomarkers in non-small-cell lung cancer can use these skills to expand a seed corpus via \texttt{citation-network}, group findings by biomarker, and produce a comparative summary table for subsequent refinement.
 
\textbf{Systematic review screening and citation management.} The \texttt{systematic-review-screener} skill applies user-defined eligibility criteria to retrieved records and produces PRISMA-compatible screening logs~\cite{page2021prisma}. Complementary AIPOCH skills include \texttt{meta-search-builder}, \texttt{\seqsplit{meta-abstract-screener}}, \texttt{meta-screening-fulltext}, and \texttt{\seqsplit{meta-criteria-generator}} for PICOS-based eligibility~\cite{aipoch_medical_skills_repo}. Reference management is handled through \texttt{citation-management}, which performs deduplication and format conversion across AMA, APA, and Vancouver styles, along with \texttt{citation-formatter}, \texttt{reference-style-sync}, and \texttt{journal-matchmaker} for venue recommendation~\cite{aipoch_medical_skills_repo}. A research group resubmitting a \seqsplit{manuscript} can convert citation styles, merge duplicates introduced during collaborative editing, and receive alternative venue suggestions in a single pipeline.
 
\textbf{Emerging topic detection.} Beyond retrieval, skills such as \texttt{emerging-topic-scout}, \texttt{research-hotspot-analysis}, \texttt{\seqsplit{keyword-velocity-tracker}}, and \texttt{translational-gap-analyzer} support the identification of growing research directions and translation gaps~\cite{aipoch_medical_skills_repo}. A junior faculty member choosing a research direction in AI for intraoperative guidance can obtain a structured briefing on high-momentum subtopics and preclinical findings still lacking clinical validation.
 
By structuring these tasks as composable agent operations, the OpenClaw and AIPOCH repositories allow a single researcher to achieve literature coverage that would previously have required a dedicated librarian or reviewer team. The outputs carry the same risks as other agent-generated content, so missed references, mischaracterized findings, and fabricated citations remain possible, and all summaries should be verified against primary sources before use in publications or clinical decisions~\cite{OpenClawMedicalSkills}.

\subsection{Clinical Documentation \& Decision Support}

\textbf{Workflow-native clinical documentation.} Recent medical AI skill collections indicate a shift from general-purpose question answering toward workflow-native clinical assistance. This shift matters because clinical notes are not merely free text. They are medico-legal, interoperable, and decision-relevant records. SOAP notes, discharge summaries, and specialist reports each serve distinct functions, yet all must preserve traceability and support continuity of care. Accordingly, an AI agent that generates documentation becomes much more useful when it can also structure that output into FHIR-compatible resources.
 
\textbf{AI-assisted note generation and safeguards.} The literature increasingly supports AI-assisted note generation and summarization, but it also emphasizes the need for strong safeguards. Clinical summarization systems may reduce documentation burden and improve synthesis of longitudinal records, yet they remain vulnerable to omission, hallucination, and factual inconsistency~\cite{oliveira2025development,chua2024integration}. These risks are especially consequential in discharge summaries, where errors may affect medication reconciliation and follow-up planning. For that reason, high-value documentation tools must be evaluated not only for fluency but also for factual grounding, source traceability, and interoperability.
 
\textbf{Standardized reporting frameworks.} Standardized reporting frameworks are equally important beyond routine notes. Their inclusion in medical agent repositories is therefore significant. It suggests that useful AI systems in medicine must be able to produce outputs that are not only readable but also compliant with accepted professional and regulatory standards.
 
\textbf{Evidence-aware decision support.} Clinical decision support in these repositories is also aligned with Grading of Recommendations, Assessment, Development and Evaluation (GRADE). The GRADE framework was developed to standardize how evidence quality and recommendation strength are represented in clinical guidance~\cite{guyatt2008grade}. For agent-based systems, this is significant because it enables recommendations to be linked to explicit evidentiary confidence rather than presented as unsupported conclusions. In this sense, the convergence of clinical documentation, FHIR integration, and GRADE-based reasoning points toward a more mature model of medical AI, centered on auditable, interoperable, and evidence-aware clinical artifacts rather than text generation alone.

\subsection{Drug Discovery \& Safety}

\textbf{Knowledge bases for drug and target annotation.} Medical AI skill collections increasingly support the full translational drug pipeline, from target and compound annotation to interaction prediction, safety surveillance, and repurposing. Foundational resources such as ChEMBL and DrugBank remain central because they provide complementary representations of drug knowledge. ChEMBL was designed as a large-scale open bioactivity resource containing binding, functional, and ADMET (absorption, distribution, metabolism, excretion, and toxicity) related data for drug-like molecules and their targets, supporting structure-activity analysis and predictive modeling~\cite{gaulton2012chembl}. DrugBank, in turn, integrates information on drugs, mechanisms, targets, and interactions, and has expanded substantially to include investigational compounds, drug-drug interactions, and repurposing-relevant data~\cite{wishart2018drugbank}. An agent capable of orchestrating both resources is therefore well suited for tasks such as target-to-compound mapping, mechanism-of-action review, off-target assessment, and candidate prioritization.
 
\textbf{Drug-drug interaction prediction.} Drug-drug interaction (DDI) prediction is one of the clearest examples of how these skill collections translate biomedical knowledge into actionable analysis. While classical DDI assessment relied on curated knowledge of CYP450 metabolism, transporters, and pharmacodynamic effects, recent work increasingly uses deep learning, graph methods, and knowledge graphs to manage the scale and complexity of polypharmacy~\cite{li2024deep}. Still, the literature emphasizes that predictive performance alone is not enough. Clinically useful systems must also preserve mechanistic interpretability. This is why skills that explicitly reason about CYP enzymes, transporters, severity, and management are more valuable than black-box prediction alone.
 
\textbf{Pharmacovigilance and safety surveillance.} Pharmacovigilance extends this logic into real-world safety monitoring. The FDA Adverse Event Reporting System (FAERS) and related spontaneous reporting systems remain essential for detecting rare and post-marketing adverse events. Established disproportionality methods such as the proportional reporting ratio (PRR) and reporting odds ratio (ROR) continue to play a central role in signal detection~\cite{fusaroli2024reporting}, and recent work increasingly combines FAERS with structured resources such as DrugBank and SIDER to improve signal prioritization and interpretation~\cite{schreier2024integration}. Repositories that expose these workflows as agent skills are therefore closely aligned with current pharmacovigilance methodology.
 
\textbf{Drug repurposing and ADMET profiling.} Drug repurposing and ADMET profiling further illustrate the value of integrated agent skill collections. Modern repurposing depends on linking disease biology, target evidence, compound bioactivity, and safety constraints into coherent hypotheses rather than relying on serendipity alone~\cite{pushpakom2019drug,morris2024drug}. At the same time, ADMET prediction remains essential because many compounds fail not for lack of activity, but because of poor absorption, metabolism, distribution, or toxicity profiles~\cite{van2003admet}. These tasks are increasingly shaped by a network pharmacology perspective, in which drugs are understood in relation to targets, pathways, and disease modules rather than single isolated interactions.

\subsection{Genomics \& Variant Interpretation}

Genomic variant interpretation is a natural fit for skill-augmented agents because it spans a long, heterogeneous chain of operations: file parsing, quality control, functional and population annotation, clinical classification, and downstream prioritization. Each step depends on different tools, reference resources, and conventions, and assembling them by hand is the main reason variant interpretation remains slow even when the underlying data are inexpensive to generate. The OpenClaw medical skills collection organizes these operations as a connected workflow rather than a collection of isolated utilities.
 
\textbf{Variant calling and file handling.} At the lowest level, skills support VCF (Variant Call Format) reading, normalization, filtering, and summary statistics over both germline and somatic call sets. Germline calling is anchored on the GATK Best Practices pipeline, which combines BWA alignment, base quality recalibration, and joint genotyping with HaplotypeCaller~\cite{vanderauwera2020gatk}, while DeepVariant offers a deep-learning alternative that achieves competitive accuracy across short-read platforms~\cite{poplin2018deepvariant}. For somatic analysis, agents wrap callers such as Mutect2 and Strelka2 to detect low-allele-fraction variants in tumor samples~\cite{kim2018strelka2}.
 
\textbf{Functional and population annotation.} Once a call set exists, the agent annotates variants against ClinVar for clinical assertions~\cite{landrum2018clinvar}, gnomAD for population allele frequencies across more than 140{,}000 individuals~\cite{karczewski2020gnomad}, dbSNP for canonical identifiers, and MyVariant.info as a unified query interface~\cite{xin2016myvariant}. Functional consequences are computed through tools such as Ensembl VEP, which assigns transcript-level effects, regulatory annotations, and pathogenicity predictions~\cite{mclaren2016vep}. The agent's role is to chain these resources so that a researcher receives an annotated table rather than a raw VCF.
 
\textbf{Clinical interpretation and prioritization.} Clinically oriented skills implement the ACMG/AMP framework for sequence variant classification, which combines criteria covering population data, computational evidence, functional studies, segregation, and de novo occurrence into a five-tier pathogenicity call~\cite{richards2015acmg}. For polygenic conditions, skills support polygenic risk score (PRS) calculation using methods such as PRSice and LDpred, allowing per-individual risk estimation from genome-wide association summary statistics~\cite{choi2020prsice,vilhjalmsson2015ldpred}. In oncology, somatic interpretation skills compute mutational signatures using the COSMIC reference catalogue~\cite{alexandrov2020cosmic} and tumor mutational burden, which has become an actionable biomarker for immune checkpoint inhibitor response~\cite{chalmers2017tmb}.
 
\textbf{Structural and copy-number variation.} Beyond single-nucleotide variants, skills support structural variant discovery through Manta and Delly for short-read data~\cite{chen2016manta,rausch2012delly} and tools such as Sniffles for long-read sequencing~\cite{sedlazeck2018sniffles}, along with copy-number and circulating tumor DNA analysis. Together, these capabilities suggest that agentic systems can support the full spectrum from raw variant data to clinically informed prioritization in rare disease, cancer genomics, and precision medicine.

\subsection{Bioinformatics Pipelines}

Beyond variant interpretation, the agentic-AI layer extends toward a workflow-oriented bioinformatics platform covering sequencing quality control, transcriptomics, epigenomics, microbiome analysis, and integrative multi-omics. Each of these areas has accumulated mature open-source tooling over the past decade, and the role of an agent is to orchestrate these tools into reproducible pipelines rather than to replace them.
 
\textbf{Quality control and bulk transcriptomics.} At the entry point, skills wrap FastQC and MultiQC for read-level quality assessment~\cite{ewels2016multiqc}, adapter trimming through Trim Galore or fastp~\cite{chen2018fastp}, and alignment through STAR or HISAT2~\cite{dobin2013star,kim2019hisat2}. For bulk RNA-seq, differential expression analysis is supported through DESeq2 and edgeR, both of which model count data with negative binomial distributions and remain the standard tools for the field~\cite{love2014deseq2,robinson2010edger}. Downstream functional interpretation uses gene set enrichment analysis and overrepresentation testing through tools such as clusterProfiler~\cite{wu2021clusterprofiler}.
 
\textbf{Single-cell and spatial omics.} Single-cell RNA-seq pipelines are anchored on Scanpy and Seurat, which provide preprocessing, normalization, dimensionality reduction, clustering, and cell-type annotation~\cite{wolf2018scanpy,hao2021seurat}. Probabilistic models such as scVI use variational autoencoders for batch correction, denoising, and integrative analysis across datasets~\cite{lopez2018scvi}. Trajectory inference is supported through methods such as PAGA and Monocle for ordering cells along developmental or differentiation axes~\cite{trapnell2014monocle}, while CellChat enables cell-cell communication analysis from ligand-receptor co-expression~\cite{jin2021cellchat}. Spatial transcriptomics adds a further layer through tools such as Squidpy, which extends single-cell analysis with spatial context~\cite{palla2022squidpy}.
 
\textbf{Epigenomics and microbiome analysis.} ATAC-seq and ChIP-seq workflows include alignment, peak calling through MACS2~\cite{zhang2008macs}, differential accessibility analysis, and motif discovery through HOMER. Higher-order chromatin organization is addressed through Hi-C analysis pipelines. Microbiome profiling is supported through QIIME~2 for amplicon sequencing and MetaPhlAn for shotgun metagenomic taxonomic profiling~\cite{bolyen2019qiime2,beghini2021metaphlan}.
 
\textbf{Multi-omics integration.} As biomedical studies increasingly generate paired transcriptomic, epigenomic, proteomic, and metabolomic data, integrative methods have become central. MOFA and its successor MOFA+ use latent factor models to identify shared and modality-specific variation across multi-omics datasets~\cite{argelaguet2020mofa}, while similarity network fusion combines patient-level networks across data types for disease subtyping~\cite{wang2014snf}. Together, these skills frame bioinformatics agents not as isolated code assistants, but as scalable computational copilots for reproducible, end-to-end analysis across bulk, single-cell, spatial, and multi-modal biological datasets.

\subsection{Protein Structure \& Design}

Within the OpenClaw medical skills collection, protein structure and design are organized as a coordinated workflow spanning structural prediction, generative design, sequence optimization, and specialized binder and antibody engineering, rather than as isolated model invocations. This framing is consistent with the broader ``Structural Biology \& Literature'' skill cluster in ClawBio, where structure-focused skills are positioned alongside tools for literature synthesis, laboratory record integration, and profile reporting. Protein engineering is operationalized as a reusable design cycle in which candidate biomolecules are proposed, structurally evaluated, contextualized against prior knowledge, and prepared for downstream validation.
 
\textbf{Structural prediction as an evaluative layer.} The central entry point is the \texttt{struct-predictor} skill, designed for local protein structure prediction using AlphaFold~\cite{jumper2021highly,abramson2024accurate}, Boltz~\cite{passaro2025boltz}, or Chai~\cite{chai2024chai}, and supporting structure comparison, RMSD calculation, and 3D visualization. Within OpenClaw, prediction is not treated as a terminal step but as part of an evaluative layer that lets agents compare candidate conformations, inspect confidence, and rank designs for further analysis. AlphaFold-family models and Boltz function as structural assessment tools for judging whether a proposed sequence is likely to realize the intended fold, or whether a candidate complex is sufficiently plausible to justify additional computational or experimental effort. They act as high-value filters that make iterative design workflows more tractable, rather than definitive proof of molecular behavior.
 
\textbf{Backbone generation and sequence assignment.} OpenClaw's protein-design capabilities align with the standard decomposition of de novo design into backbone generation and sequence assignment. RFdiffusion~\cite{watson2023novo} occupies the backbone-generation role, sampling candidate scaffolds under specified geometric or functional constraints and enabling exploration of folds and interfaces beyond naturally occurring templates. ProteinMPNN~\cite{sumida2024improving} addresses the inverse-folding problem by mapping candidate backbones into sequence space, proposing amino acid sequences likely to stabilize the generated structures. Together they form the core generative engine: RFdiffusion expands the structural search space, while ProteinMPNN converts promising topologies into concrete molecular candidates. In an agentic setting, this division supports staged decision-making, retaining structural diversity early and committing resources to sequence-level hypotheses only later.
 
\textbf{Binder and antibody engineering.} BindCraft~\cite{pacesa2024bindcraft} functions as a campaign-level orchestration layer that integrates backbone generation, interface optimization, sequence refinement, and structure-based filtering into a single automated workflow. Binder discovery becomes a composable search process rather than a manually assembled series of scripts, consistent with the broader skill-based design philosophy of the repository. Antibody engineering extends the same logic into a therapeutically specialized setting. Although antibodies share the binder-design objective of high specificity and affinity, they introduce additional constraints related to complementarity-determining region (CDR) flexibility, epitope targeting, humanization, stability, and developability~\cite{raja2025next,kandari2023antibody}. Antibody-oriented skills are therefore better framed as domain-adapted workflows that place greater weight on loop-level optimization and candidate prioritization under therapeutic constraints.
 
\textbf{Integration with the wider research stack.} OpenClaw situates protein design within a larger research-support infrastructure. The \texttt{lit-synthesizer} skill searches PubMed and bioRxiv, summarizes papers, constructs citation graphs, and generates review sections, linking structural design workflows directly to current evidence and prior art. The \texttt{labstep} skill provides access to the Labstep electronic lab notebook API for querying experiments, protocols, and inventory, bridging in silico design and laboratory execution, while reporting skills such as \texttt{profile-report} support structured synthesis of results. Taken together, OpenClaw does not conceptualize protein structure and design as a narrow prediction task, but as one component of a broader agent-assisted research workflow that connects computation, literature, and experimentation.

 
\subsection{Medical Imaging \& Pathology}
\label{sec:imaging}

Medical imaging and pathology are among the more mature application domains for agentic AI, and the OpenClaw medical skills collection reflects this through capabilities spanning image handling, computational pathology, multimodal understanding, and biosignal processing. Representative skills include \texttt{pydicom} for DICOM reading, windowing, and format conversion; \texttt{histolab} and \texttt{pathml} for whole-slide image tiling, stain normalization, and patch extraction; \texttt{neurokit2} for ECG, EEG, and respiration feature extraction; and \texttt{computational-pathology-agent} and \texttt{\seqsplit{multimodal-medical-imaging}} for integrated imaging analysis. Together they support end-to-end imaging workflows integrating acquisition, preprocessing, interpretation, and report generation within a single composable pipeline.
 
\textbf{DICOM handling and preprocessing.} The \texttt{pydicom} skill provides programmatic access to DICOM files, including pixel arrays, metadata headers, modality-specific windowing conventions, and format conversion. A radiologist applying a custom deep learning model to a batch of chest CT studies exported from a PACS can use it to iterate over the study directory, extract axial slices at the appropriate window-level preset, and write normalized NIfTI volumes for downstream inference. Because DICOM handling is a prerequisite for nearly every imaging task, it functions as a foundational layer that lets higher-level skills receive well-formed inputs without managing format-specific parsing.
 
\textbf{Whole-slide imaging and computational pathology.} The central challenge in computational pathology is scale: a single whole-slide image (WSI) can exceed $100{,}000 \times 100{,}000$ pixels, making direct inference infeasible without principled tiling. The \texttt{histolab} and \texttt{pathml} skills encapsulate patch-based pipelines that tile slides at configurable magnification, filter uninformative background, apply Macenko or Vahadane stain normalization to reduce cross-institutional batch effects, and pass patches to downstream classification or feature extraction models. An example is a translational oncology researcher quantifying tumor-infiltrating lymphocyte (TIL) density across 200 breast cancer WSIs from two hospitals with different staining protocols: the agent extracts and normalizes patches, calls a pretrained TIL detection model wrapped in a second skill, and aggregates per-patch predictions into a per-slide score written to CSV for survival analysis. This aligns with broader work on AGI for medical image analysis, which stresses preprocessing standardization before model inference~\cite{li_2024_agi_imaging}. Recent work on pathology foundation models suggests that future WSI skills will incorporate contrastive pretrained encoders queryable through natural language rather than fixed task-specific heads~\cite{Lu2024CONCH}.
 
\textbf{Cardiac imaging and echocardiography.} Large-scale self-supervised pretraining on echocardiography videos can produce spatiotemporal cardiac representations that generalize across view identification, chamber segmentation, and disease classification~\cite{kim_2025_echofm}. In an agentic setting, a cardiologist screening transthoracic echocardiograms for reduced ejection fraction before clinical trial enrollment could invoke a cardiac imaging skill to identify the apical four-chamber view, segment the left ventricular cavity across the cardiac cycle, and compute the biplane Simpson ejection fraction, producing a structured table with per-study confidence scores and flagging sub-threshold studies for manual review. While the OpenClaw collection does not yet include a dedicated echocardiography skill, the \texttt{multimodal-medical-imaging} skill provides a general-purpose foundation for such workflows.
 
\textbf{Segmentation and biosignal analysis.} Segment Anything and its biomedical adaptations generalize across imaging modalities and anatomical targets given appropriate prompts~\cite{yan_2024_biomedical_sam2}. In a segmentation workflow, the agent accepts a bounding box, a point click, or a text description and produces a binary mask for downstream volumetric analysis. An abdominal radiologist measuring hepatic lesion volume across three follow-up CT timepoints can segment each study via a point prompt at the lesion center, compute volumetrics, and return a RECIST 1.1 longitudinal summary. Biosignal analysis is supported by the \texttt{neurokit2} skill, which provides preprocessing and feature extraction for physiological time series including ECG, EEG, EDA, RSP, PPG, and EMG signals. A sleep researcher can extract heart rate variability metrics and sleep-stage-stratified respiratory patterns from overnight polysomnography across 50 participants and export a tidy data frame for statistical modeling.
 
These capabilities frame the agent as an imaging workflow orchestrator, coordinating DICOM access, \seqsplit{task-appropriate preprocessing}, foundation-model inference, and structured reporting. Whether this promise is realized in practice depends on skill reliability, underlying model accuracy, and orchestration correctness. These challenges are discussed in Section~\ref{sec:challenges}.

 
\subsection{Regulatory \& Medical Devices}
\label{sec:regulatory}

The regulatory domain presents some of the most complex challenges for agent-assisted biomedical research, and the OpenClaw medical skills collection addresses it through a set of capabilities for compliance workflows, documentation preparation, and risk analysis within existing frameworks. Representative skills include \texttt{\seqsplit{regulatory-drafter}} (510(k) and premarket approval (PMA) document drafting with predicate device comparison tables), \texttt{\seqsplit{regulatory-drafting}} (EU MDR and IVDR technical documentation and clinical evaluation report templates), \texttt{iso-13485-certification} (quality management system documentation per ISO~13485), \texttt{\seqsplit{clinical-trial-protocol-skill}} (ICH-aligned clinical trial protocol drafting), \texttt{fda-database} (querying the FDA device and drug databases), and \texttt{fda-guideline-search} (retrieval of FDA guidance documents)~\cite{OpenClawMedicalSkills,aipoch_medical_skills_repo}. These skills do not replace regulatory expertise or formal compliance processes; they assist researchers and regulatory affairs professionals in working through procedural and documentation requirements.
 
\textbf{FDA and CE mark compliance.} In the United States, AI-enabled medical devices enter the market primarily through FDA authorization, which by the end of 2025 covered over 1{,}400 AI and machine learning-enabled devices~\cite{FDATracker2025}. Devices must demonstrate analytical and clinical validation, establish a predetermined change control plan, and meet Quality System Regulation requirements. The \texttt{regulatory-drafter} skill assists with 510(k) preparation, predicate comparison summaries, and substantial equivalence templates. A medical device startup developing an AI-based retinal image grader for diabetic retinopathy screening can have the agent identify a suitable predicate from the 510(k) database, draft a comparison table covering intended use and performance characteristics, and populate a shell submission document. The \texttt{regulatory-drafting} skill supports complementary CE marking workflows under the European MDR and IVDR, including technical documentation, clinical evaluation reports, and declaration of conformity drafting. These capabilities are especially relevant given the EU AI Act timeline, which classifies AI systems embedded in medical devices as high-risk and mandates full compliance by August 2027~\cite{EUAIAct2024}.
 
\textbf{Quality management and software lifecycle.} Medical device quality management is addressed through the \texttt{iso-13485-certification} skill, which supports documentation of quality management system requirements per \seqsplit{ISO~13485}, including design controls, verification and validation records, and traceability matrices. For software-specific standards such as IEC~62304, the skill helps development teams classify components by safety class (Class~A, B, or C, depending on severity of potential harm from software failure) and document the required activities and maintenance procedures. A hospital informatics team building an ML-based sepsis risk tool can have the risk-scoring module classified as Class~C (injury or death possible on failure), with a generated documentation checklist covering design inputs, software requirements, architecture, unit and integration testing, and a traceability matrix linking requirements to verification evidence. For risk management per ISO~14971, agents can support structured risk assessment by walking through hazard identification for each use scenario, estimating severity and probability of harm, applying manufacturer-defined acceptability criteria, and producing a risk control table documenting mitigations and residual risk ratings.
 
\textbf{Clinical trial protocol generation.} The \texttt{\seqsplit{clinical-trial-protocol-skill}} drafts protocol elements required by ICH guidelines, including study objectives, eligibility criteria, endpoints, statistical analysis plans, safety monitoring provisions, and data management procedures. A trialist designing a Phase II adaptive trial of a novel anti-fibrotic agent in idiopathic pulmonary fibrosis can iteratively refine a draft shell that includes an adaptive sample size re-estimation rule based on an interim analysis of forced vital capacity decline, pre-specified stopping rules for futility and efficacy, an eligibility checklist, and a statistical analysis plan covering the primary analysis, sensitivity analyses, and handling of missing data.
 
Regulatory documentation carries legal and compliance consequences, and errors in a 510(k) submission or a risk management file can delay clearance or expose manufacturers to liability. Agent-generated regulatory documents therefore require rigorous human review by qualified personnel before submission, and should be treated as structured drafts rather than authoritative outputs. The value of agent-assisted regulatory workflows lies in reducing the effort required to produce high-quality first drafts that experts can then review and refine.

\subsection{Health \& Wellness}
\label{sec:health}

Health and wellness extend agent assistance beyond diagnosis and treatment into prevention, self-management, and longitudinal behavior change. These capabilities are well-suited to repeated, low-stakes interactions in which behavioral data, self-reports, and wearable signals can be iteratively interpreted over time~\cite{OpenClawMedicalSkills,aipoch_medical_skills_repo}. Representative skills include \texttt{nutrition-analyzer}, \texttt{sleep-analyzer}, \texttt{fitness-analyzer}, \texttt{tcm-constitution-analyzer}, and \texttt{mental-health-analyzer}, \texttt{\seqsplit{wearable-analysis-agent}}, \texttt{goal-analyzer}, \texttt{\seqsplit{health-trend-analyzer}}, \texttt{weightloss-analyzer}, and crisis-oriented escalation skills~\cite{OpenClawMedicalSkills}.
 
\textbf{Nutrition and dietary planning.} WHO emphasizes that a healthy diet helps protect against malnutrition and noncommunicable diseases, including diabetes, cardiovascular disease, and some cancers~\cite{who_healthy_diet_2026}. The \texttt{nutrition-analyzer} skill supports macro- and micronutrient tracking, dietary assessment, meal planning, and food data lookup, moving the agent from passive logging to structured nutritional reasoning~\cite{OpenClawMedicalSkills}. A patient with obesity and early insulin resistance can upload a week of meal logs and weight records; the agent, together with \texttt{weightloss-analyzer}, estimates caloric balance, flags excessive refined carbohydrate intake or low fiber, and generates a revised meal plan organized around improved satiety, lower glycemic load, and more consistent protein distribution.
 
\textbf{Sleep and circadian health.} A joint consensus statement from the American Academy of Sleep Medicine and the Sleep Research Society recommends at least 7 hours of sleep per night for healthy adults~\cite{watson2015sleep}. The \texttt{sleep-analyzer} skill interprets sleep duration, efficiency, staging, and circadian patterns, while \texttt{wearable-analysis-agent} incorporates activity, heart rate variability, and ECG streams~\cite{OpenClawMedicalSkills}. A medical student reporting chronic fatigue, fragmented sleep, and impaired daytime concentration can receive a structured interpretation pointing to circadian misalignment and insufficient sleep opportunity, along with recommendations such as fixed wake times, morning light exposure, and reduced evening stimulant intake.
 
\textbf{Fitness and rehabilitation.} WHO recommends at least 150 minutes of moderate-intensity activity per week, or 75 minutes of vigorous-intensity activity, plus muscle-strengthening on two or more days~\cite{who_physical_activity_2024}. The \texttt{\seqsplit{fitness-analyzer}} skill supports exercise tracking, strength and cardiorespiratory performance assessment, training load estimation, and periodization, while \texttt{rehabilitation-analyzer} extends this logic to functional recovery~\cite{OpenClawMedicalSkills}. An individual resuming exercise after a knee injury can provide step counts, range-of-motion notes, and training logs; the agent estimates workload tolerance, distinguishes expected soreness from possible overload, and generates a phased return-to-training plan with mobility work, strength progression, and recovery checkpoints.
 
\textbf{Traditional Chinese Medicine.} WHO adopted the \seqsplit{Global} Traditional Medicine Strategy 2025--2034, which emphasizes strengthening evidence, ensuring safety, and integrating traditional, complementary, and integrative medicine into health systems where appropriate~\cite{who_traditional_medicine_2025}. The \texttt{\seqsplit{tcm-constitution-analyzer}} skill supports constitution assessment, pattern differentiation, and lifestyle-oriented recommendations based on user-reported symptoms~\cite{OpenClawMedicalSkills}. A user reporting chronic fatigue, poor appetite, cold intolerance, and non-restorative sleep can have these observations organized into a TCM-style profile with candidate syndrome patterns and general recommendations on routine, diet, and broad herbal categories for practitioner discussion. Such outputs should not substitute for evaluation by a licensed TCM practitioner.
 
\textbf{Mental health analytics.} WHO emphasizes that mental health is an integral component of health and that promotion, prevention, care, and recovery require sustained system- and community-level support~\cite{who_mental_health_2025}. The \texttt{mental-health-analyzer} skill supports mood tracking, symptom-pattern analysis, and PHQ/GAD-style screening logic, consistent with the widespread use of the Patient Health Questionnaire (PHQ-9) for depression severity and the Generalized Anxiety Disorder scale (GAD-7) for anxiety severity~\cite{kroenke2001phq9,spitzer2006gad7}. Related crisis-oriented skills support escalation when higher-risk signals appear~\cite{OpenClawMedicalSkills}. A university student maintaining daily mood logs, sleep records, and journal entries can receive a structured summary indicating possible deterioration in well-being, with recommendations for closer monitoring, coping strategies, or professional follow-up, while higher-risk cases are routed to crisis-aware protocols. An important design principle is that low-stakes wellness analytics and high-stakes crisis handling should remain architecturally distinct even when they rely on overlapping signals.
 
Wellness skills shift medical agents from episodic interaction toward continuous, preventive support, since nutrition, sleep, fitness, TCM self-care, and mental health tracking all generate longitudinal data compatible with agentic synthesis. These domains remain medically consequential: poor recommendations about diet, exercise, herbal products, or mental health risk can still cause harm, particularly for users with chronic disease, psychiatric vulnerability, or polypharmacy. Outputs should therefore be treated as structured guidance and first-pass analysis rather than definitive medical advice, helping users and professionals prioritize follow-up and sustain healthier behavior under appropriate human oversight.

\subsection{Data Science \& Scientific Computing}

Data science and scientific computing constitute the general methodological layer of medical agents, providing shared computational foundations across research settings: statistical analysis, survival modeling, Bayesian inference, scientific visualization, and laboratory automation. As biomedical agents move beyond knowledge-based question answering toward workflow execution, data preparation, quantitative analysis, result presentation, and tool use have become central to their practical utility~\cite{XuSankar2025,Njeietal2026}. This layer should therefore be understood as a prerequisite for integrating agents into real research workflows.
 
\textbf{Statistical analysis.} Much of the work in clinical and experimental research lies in data cleaning, variable encoding, outlier handling, group comparison, and baseline table construction rather than in complex models themselves. Without these steps, even a linguistically capable agent cannot reliably produce reviewable quantitative conclusions. Systems such as ToolUniverse already integrate tool invocation and code execution into unified research pipelines, turning the agent into part of an executable analytical process rather than a natural-language interface alone~\cite{Gaoetal2025ToolUniverse}. A representative scenario is clinical cohort analysis, where demographic variables, laboratory measurements, and outcomes must be compared across exposure groups; here the value of the agent lies in workflow integration, traceability, and error reduction rather than in replacing statistical judgment.
 
\textbf{Survival modeling.} Time-to-event data are common in oncology, cardiovascular studies, critical care, and long-term follow-up research, where the key methodological challenges include censoring, dynamic risk sets, and multivariable prognostic modeling. The Kaplan--Meier estimator and the Cox proportional hazards model remain the classical foundations for nonparametric survival estimation and semiparametric risk modeling~\cite{KaplanMeier1958,Cox1972}. A translational oncology researcher evaluating the prognostic value of a candidate gene alongside clinical covariates benefits less from raw result generation than from a system that organizes the full analytical workflow, covering risk stratification, covariate adjustment, censoring-aware evaluation, and figure preparation.
 
\textbf{Bayesian optimization and experimental design.} In many life science experiments, high cost, long iteration cycles, and large parameter spaces make the question of what experiment to run next a computational problem in its own right. Bayesian optimization identifies the most informative parameter combinations under constrained experimental budgets by modeling both the objective function and its uncertainty~\cite{Snoeketal2012}. This is especially relevant to self-driving laboratories, culture-condition screening, and reagent formulation. Tool-augmented scientific agents such as ChemCrow have shown that combining language reasoning with specialized computational tools can substantially expand the action space of agents in scientific tasks~\cite{Branetal2024ChemCrow}.
 
\textbf{Scientific visualization.} Figures are part of scientific argumentation, since axis scaling, error representation, color encoding, and layout all affect interpretability. Matplotlib has long served as a foundational environment for reproducible publication-quality graphics~\cite{Hunter2007}. Within medical-agent workflows, this extends to automated organization of multi-panel figures, survival plots, heatmaps, and comparative visualizations. In manuscript figure preparation, where expression differences, survival analyses, and correlation structures must be assembled into a unified layout, an agent that coordinates both analysis and plotting supports the standardization of scientific communication.
 
\textbf{Data management and laboratory automation.} Scientific computing also covers large-scale data object management, time-series analysis, and laboratory automation. With single-cell sequencing, spatial transcriptomics, and multimodal data integration, biomedical research increasingly depends on standardized data containers and reusable pipelines. Laboratory automation is beginning to push this layer from analysis toward execution: the Opentrons Python Protocol API makes liquid-handling procedures explicitly programmable~\cite{OpentronsDocs}, and PyLabRobot provides a hardware-agnostic interface through which liquid-handling robots and accessories can be controlled via a unified Python abstraction, improving method sharing across platforms~\cite{Wierengaetal2023}. The data-science capabilities of medical agents are therefore no longer confined to post hoc analysis, but are increasingly linked to automated experimental execution, bringing analysis, decision-making, and laboratory action into a more complete closed loop.

\section{Case Studies: Vibe Medicine in Action}
\label{sec:case}

The following repository skills were inspected and used:
\begin{itemize}
\item \texttt{tooluniverse-rare-disease-diagnosis}
\item \texttt{tooluniverse-variant-interpretation}
\item \texttt{medical-research-toolkit}
\item \texttt{tooluniverse-gwas-drug-discovery}
\item \texttt{tooluniverse-drug-drug-interaction}
\item \texttt{tooluniverse-clinical-trial-matching}
\item \texttt{tooluniverse-single-cell}
\item \texttt{tooluniverse-clinical-trial-design}
\item \texttt{clinical-trial-protocol-skill}
\end{itemize}

\subsection{Case Study 1: From Phenotype to Diagnosis}

\begin{figure*}[t]
    \centering
    \includegraphics[width=0.9\textwidth]{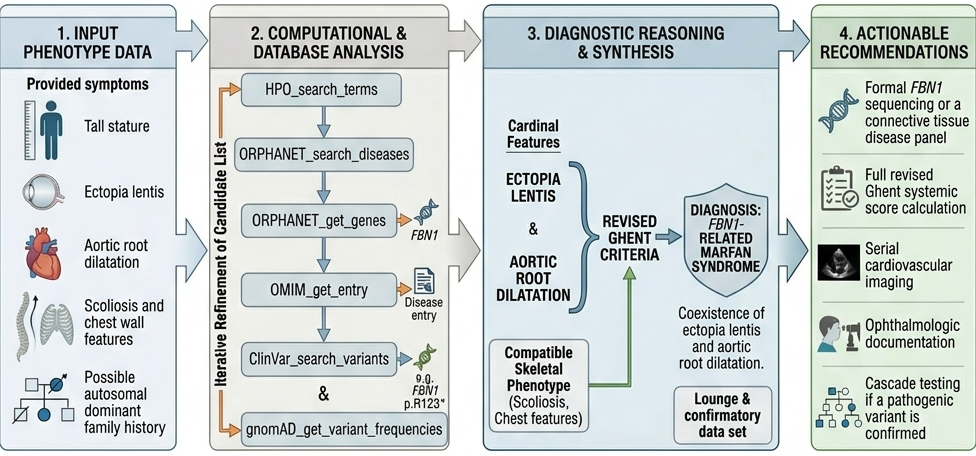}
    \caption{Diagnostic workflow for \textit{FBN1}-related Marfan syndrome. The pipeline integrates Human Phenotype Ontology (HPO) phenotype mapping with Orphanet/OMIM searches and ClinVar variant interpretation to confirm clinical findings and prioritize genetic testing.}
    \label{fig:case_study_1}
\end{figure*}

This case study was based primarily on the \texttt{\seqsplit{tooluniverse-rare-disease-diagnosis}} and \texttt{\seqsplit{tooluniverse-variant-interpretation}} skills. As shown in Fig. \ref{fig:case_study_1}, the documented workflow in the repository proceeds through symptom intake, Human Phenotype Ontology (HPO) term matching, rare disease search across Orphanet and OMIM, candidate gene prioritization, ClinVar/gnomAD-backed variant interpretation, and treatment recommendation.

A representative case was selected with the following phenotype:
\begin{itemize}
\item tall stature,
\item ectopia lentis,
\item aortic root dilatation,
\item scoliosis and chest wall features,
\item possible autosomal dominant family history.
\end{itemize}
This example was chosen because the repository itself repeatedly uses Marfan syndrome as a canonical rare-disease diagnosis scenario.

The workflow invokes the following ToolUniverse tools exposed by the skill:
\begin{itemize}
\item \texttt{HPO\_search\_terms}
\item \texttt{Orphanet\_search\_diseases}
\item \texttt{Orphanet\_get\_genes}
\item \texttt{OMIM\_get\_entry}
\item \texttt{ClinVar\_search\_variants}
\item \texttt{gnomAD\_get\_variant\_frequencies}
\end{itemize}

The phenotype mapped most strongly to \textbf{\textit{FBN1}-related Marfan syndrome}. The principal reasons were the coexistence of ectopia lentis and aortic root dilatation, which the revised Ghent criteria treat as cardinal features, together with a compatible skeletal phenotype. The top recommended next steps were:
\begin{itemize}
\item formal \textit{FBN1} sequencing or a connective tissue disease panel,
\item full revised Ghent systemic score calculation,
\item serial cardiovascular imaging,
\item ophthalmologic documentation, and
\item cascade testing if a pathogenic variant is confirmed.
\end{itemize}

\subsection{Case Study 2: Drug Repurposing Pipeline}

\begin{figure*}[t]
    \centering
    \includegraphics[width=0.9\textwidth]{Figures/case_study_2.png}
    \caption{Drug repurposing pipeline for Myasthenia Gravis. The execution highlights disease resolution via OpenTargets, survey of active trials for zilucoplan, and safety label interrogation for eculizumab using the OpenFDA tool branch.}
    \label{fig:case_study_2}
\end{figure*}

This case study followed the \texttt{\seqsplit{medical-research-toolkit}}, \texttt{\seqsplit{tooluniverse-gwas-drug-discovery}}, and \texttt{\seqsplit{tooluniverse-drug-drug-interaction}} skills. The repository workflow for drug repurposing was to identify disease-associated targets, assess target tractability, retrieve compounds, evaluate safety and drug-drug interactions, and survey recent and active clinical trials, as depicted in Fig. \ref{fig:case_study_2}.

The disease chosen for the live example was \textbf{myasthenia gravis}. This choice was deliberate because the repository documentation itself repeatedly uses myasthenia gravis in the \texttt{medical-research-toolkit} examples and the public MCP endpoint was demonstrably able to resolve it through OpenTargets.

The public MCP endpoint described by the repository documentation was used as the main live backend:
\begin{itemize}
    \item \texttt{opentargets\_search} to resolve the disease,
    \item \texttt{\seqsplit{opentargets\_get\_associations}} to retrieve disease-associated targets,
    \item \texttt{ctg\_search\_studies} to survey recent and active trials,
    \item \texttt{openfda\_search\_drug\_labels} to review FDA-labeled warnings for a candidate drug,
    \item \texttt{pubmed\_search\_articles} to obtain supporting treatment literature.
\end{itemize}

Using \texttt{opentargets\_search} on ``myasthenia gravis'' returned the disease entry \texttt{EFO\_0004991}, named Myasthenia gravis, with a description identifying it as a rare autoimmune disorder of the neuromuscular junction characterized by fatigable weakness. This provided a concrete disease identifier for downstream reasoning.

Using \texttt{ctg\_search\_studies} for recruiting myasthenia \seqsplit{gravis} trials returned active or recent study records in the disease space. One surfaced example in the session included an extension study of zilucoplan in generalized myasthenia gravis. This established that the trial-matching branch of the repository workflow was live for the chosen disease.

The OpenFDA tool branch was used to query labeling information for \textbf{eculizumab}, a clinically relevant complement inhibitor in generalized myasthenia gravis. The tool returned FDA label content including a boxed warning for serious meningococcal infections and the requirement for vaccination and REMS-based safeguards. This was a concrete example of how the repository's safety review step can be run against a real candidate.

A PubMed search for myasthenia gravis treatment returned recent literature records, including a real-world Italian study on prevalence, treatment patterns, and economic burden. This step was used not to identify mechanistic novelty, but to verify that the literature branch of the workflow was accessible and disease-relevant.

The strongest evidence-backed interpretation from the executed steps was that myasthenia gravis can be analyzed through the public MCP endpoint in a way that meaningfully reproduces the repository workflow:
\begin{itemize}
    \item disease resolution succeeded,
    \item trial search succeeded,
    \item safety label interrogation succeeded,
    \item treatment literature retrieval succeeded.
\end{itemize}

\subsection{Case Study 3: Clinical Trial Design}

\begin{figure*}[t]
    \centering
    \includegraphics[width=0.9\textwidth]{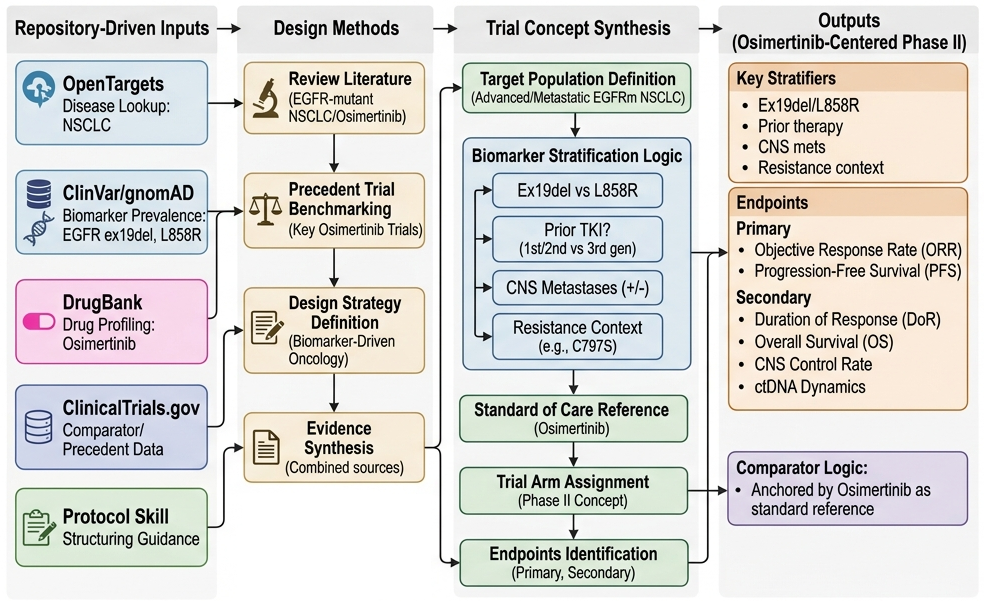}
    \caption{Clinical trial design workflow for EGFR-mutant NSCLC. The schematic shows the synthesis of biomarker prevalence and precedent trial benchmarking to generate an osimertinib-centered Phase II concept with stratified endpoints.}
    \label{fig:case_study_3}
\end{figure*}

This case study was based on the \texttt{\seqsplit{tooluniverse-clinical-trial-design}}, \texttt{\seqsplit{tooluniverse-clinical-trial-matching}}, and \texttt{\seqsplit{clinical-trial-protocol-skill}} workflows. As shown in Fig. \ref{fig:case_study_3}, the repository frames clinical trial design as an evidence-driven process involving disease characterization, biomarker prevalence, comparator profiling, precedent trial benchmarking, eligibility logic, and protocol drafting.

The trial-design example used \textbf{Epidermal Growth Factor Receptor (EGFR) mutant non-small-cell lung cancer (NSCLC)} with an \textbf{osimertinib-centered Phase II concept}. This choice matched both the repository examples and the underlying trial-design logic, which is strongest in biomarker-defined oncology.

Under ideal conditions, the repository workflow would invoke tools for:
\begin{itemize}
\item OpenTargets disease lookup,
\item ClinVar/gnomAD biomarker prevalence checks,
\item DrugBank drug profiling,
\item ClinicalTrials.gov retrieval,
\item and protocol-template generation.
\end{itemize}

The design exercise was carried out by combining:
\begin{itemize}
\item repository workflow instructions,
\item review literature on EGFR-mutant NSCLC and osimertinib,
\item surfaced ClinicalTrials.gov examples in the osimertinib space,
\item and protocol-structuring guidance from the installed protocol skill.
\end{itemize}

The executed output was a biomarker-stratified Phase II concept with:
\begin{itemize}
\item an EGFR-mutant advanced NSCLC population,
\item key stratifiers including exon 19 deletion, L858R, prior therapy, central nervous system (CNS) metastases, and resistance context,
\item candidate primary endpoints of objective response rate or progression-free survival,
\item secondary endpoints including duration of response, overall survival, CNS control, and ctDNA dynamics,
\item a realistic comparator logic anchored around osimertinib as standard reference therapy.
\end{itemize}

\section{Challenges and Risks}
\label{sec:challenges}

While Vibe Medicine offers substantial promise for democratizing biomedical computation, its current form presents a number of significant challenges spanning technical, legal, ethical, and practical dimensions that must be resolved before the paradigm can be adopted responsibly in clinical and research settings.
 
\textbf{Hallucination and Misinformation.}
The most immediate risk is factual error. LLMs generate plausible but incorrect outputs, a phenomenon known as hallucination, and this tendency persists when a model operates as an agent \cite{Salehietal2025}. In a Vibe Medicine workflow, hallucination can manifest at multiple levels: fabricated citations from a literature skill, incorrect pathogenicity classifications from a variant interpretation skill, or internally coherent but factually wrong clinical reports. The consequences differ qualitatively from those in general software, because biomedical errors can affect patient safety and research validity. Grounding outputs in verifiable sources (PubMed, ClinVar, ClinicalTrials.gov) reduces but does not eliminate this risk, since the agent may still misinterpret retrieved data. No reliable automated method currently detects all forms of biomedical hallucination, making domain expertise an irreducible requirement.
 
\textbf{Data Privacy and HIPAA Compliance.}
Many biomedical tasks involve protected health information (PHI) subject to HIPAA and GDPR. Agent workflows introduce several privacy risks: cloud-hosted models may transmit patient data to external servers; persistent conversation logs may store PHI in unencrypted files; and skills querying external APIs may inadvertently include patient-identifying information in outbound requests. OpenClaw and NanoClaw mitigate these risks through local execution and ephemeral container isolation \cite{NanoClawDev}, but the underlying LLM API call still requires data to leave the local machine. Formal compliance auditing and institutional review remain necessary for any clinical deployment.
 
\textbf{Reproducibility.}
LLM outputs are inherently \seqsplit{non-deterministic}: identical prompts can yield different code, analysis paths, or conclusions across runs. Agent behavior also depends on installed skill versions and the state of external APIs at execution time, and conversational workflows are difficult to capture in a form others can inspect. Mitigations include comprehensive session logging, skill versioning tied to specific runs, deterministic execution modes where available, and publication of full interaction transcripts as supplementary material. Standardized reporting guidelines for agent-assisted research would further support reproducibility.
 
\textbf{Regulatory and Legal Considerations.}
The regulatory status of agent-generated biomedical outputs is largely undefined. It is unclear whether a Vibe Medicine output (a clinical report, a drug interaction analysis, a treatment recommendation) constitutes a medical device, a clinical decision support tool, or a research artifact. The FDA's lifecycle-based oversight frameworks were designed for narrowly scoped validated algorithms, not general-purpose agents composing pipelines from hundreds of skills \cite{FDATracker2025}. The EU AI Act classifies AI in medical devices as high-risk with compliance required by August 2027 \cite{EUAIAct2024}. Liability allocation among the researcher, framework developer, skill author, and model provider when an agent-generated analysis contributes to a clinical error remains legally untested.
 
\textbf{Security.}
Agents that execute code and query external APIs introduce a substantial attack surface. Prompt injection, in which malicious instructions embedded in retrieved content cause unintended agent actions, is the most widely cited threat and is considered unsolved in current LLM architectures \cite{OWASP2025}. The open-source, community-contributed skill collection also creates supply chain risk: security researchers have demonstrated that third-party OpenClaw skills can perform data exfiltration without user awareness. Skill vetting, provenance verification, and runtime monitoring are essential components of responsible deployment.
 
\textbf{The ``Vibe Medicine Hangover.''}
Perhaps the most insidious risk is epistemological: repeated acceptance of polished agent outputs without critical scrutiny erodes the user's own analytical capacity. This pattern has already been observed in Vibe Coding, where developers accumulate technical debt by accepting AI-generated code they do not fully understand \cite{MooreTatonetti2025}. In medicine the stakes are higher, because a clinician may lack the specialized knowledge to detect subtle errors in agent reasoning. We term this the \textbf{Vibe Medicine Hangover}: ease of use traded silently for depth of understanding. Mitigation requires mandatory human review of all agent outputs, training that emphasizes critical evaluation, and workflow designs that expose rather than hide the agent's reasoning chain.
 
\textbf{Trustworthiness.}
Trustworthiness in LLMs is multi-dimensional, encompassing truthfulness, calibration, fairness, privacy, and accountability, and cannot be reduced to task accuracy alone \cite{huang2024trustllm,Ferdaus2026TrustworthyAI,Huang2024VVSurvey}. Apparent reasoning fluency does not imply faithful scientific reasoning: models may rely on pattern completion, brittle heuristics, or artifacts in training data. In scientific domains where ground truth is incomplete and benchmarks are sparse, these limitations are amplified. Trustworthy agent outputs therefore require assessment at the model level (calibration, robustness), the data level (provenance, confounding control), the workflow level (reproducible pipelines, logging), and the scientific level (external validation and experimental replication). Generative AI in biomedical research should be treated as a hypothesis-generating instrument whose outputs require verification before acceptance as discovery.

\section{Future Directions}
\label{sec:future}

The Vibe Medicine paradigm described in this paper represents an early stage in a longer trajectory. Several directions of development are likely to shape the field in the coming years.
 
\textbf{Multi-Agent Collaboration for Clinical Workflows.}
Current workflows are orchestrated by a single agent invoking skills sequentially. A natural extension is multi-agent collaboration, in which specialized agents, each configured with a distinct subset of skills and domain expertise, work in parallel or in dialogue to address complex clinical problems. A radiogenomics investigation might involve a medical imaging agent extracting radiomic features, a genomics agent performing variant calling and pathway analysis, a literature agent retrieving and synthesizing relevant publications, and a clinical agent integrating findings into a structured report. This architecture mirrors the structure of multidisciplinary clinical teams, such as tumor boards, in which specialists contribute complementary perspectives to a shared decision. Frameworks such as CAMEL and MetaGPT have explored multi-agent coordination in general settings \cite{Lietal2023CAMEL,Hongetal2024MetaGPT}, but their adaptation to clinical workflows, where errors carry real consequences and outputs must be traceable, remains an open research problem.
 
\textbf{Closed-Loop Integration with Clinical Infrastructure.}
For Vibe Medicine to move beyond research prototyping and into clinical practice, agents must interface directly with institutional data systems: electronic health records (EHR) via Fast Healthcare Interoperability Resources (FHIR), picture archiving and communication systems (PACS) via DICOMweb, and laboratory information management systems (LIMS). OpenClaw medical skills already includes skills for FHIR development, DICOM handling, and clinical report generation, but institutional deployment will additionally require auditable authentication, access control, and compliance with local data governance policies, requirements that are well understood in health IT but have not yet been incorporated into agent framework designs. NanoClaw's container isolation model offers a promising starting point, but substantial engineering work remains before agent-EHR integration meets production clinical standards.
 
\textbf{Personalized Skill Recommendations.}
A 1,000-skill collection is too large for any individual user to survey manually, and effective skill selection depends on skill description quality, request specificity, and the agent's understanding of research context. Future systems may benefit from personalized recommendation engines that learn from a researcher's prior interactions, publication history, and institutional affiliations to suggest relevant skills proactively. Such systems could also detect gaps in a workflow, for example identifying that a genomic analysis lacks a downstream pathway enrichment step, and recommend additional skills accordingly.
 
\textbf{Self-Improving Agent Architectures.}
The skills described in this paper are static: each \texttt{SKILL.md} is written by a human developer, and the agent applies it as-is across sessions. A natural evolution is toward agents that create, refine, and optimize their own skills from experience. Hermes Agent, developed by Nous Research, represents an early implementation of this idea~\cite{HermesAgent2026}. Hermes includes a closed learning loop in which the agent automatically generates a skill document after completing a task, retrieves it when a similar task recurs, and refines it over successive executions. The companion self-evolution system applies GEPA (Generic Evolution of Prompt Architectures), which reads full execution traces (not only pass/fail outcomes) to diagnose inefficiencies and propose targeted skill mutations using DSPy-based genetic optimization. Hermes is model-agnostic (supporting over 200 models through OpenRouter, local inference, and custom endpoints), natively compatible with OpenClaw skills through a built-in migration tool, and offers a Singularity backend for deployment on shared HPC clusters common in academic biomedical computing. For Vibe Medicine, self-improving agents could enable skills that adapt to institutional analysis conventions, learn optimal parameter defaults from repeated genomic or imaging workflows, and reduce the manual skill-engineering burden that currently limits the pace at which the collection grows. Whether this closed-loop approach can maintain safety and reproducibility in regulated medical contexts, where deterministic behavior and auditability are expected, remains an open question that will require careful evaluation.

\textbf{Federated and Institutional Skill Development.}
The current OpenClaw medical skills collection is developed by a centralized open-source community. An important future direction is federated skill development, in which individual institutions create and share skills encoding their own protocols, formularies, imaging workflows, or regulatory requirements. A cancer center might contribute skills tailored to its molecular tumor board process; a genomics core facility might contribute skills for its preferred sequencing pipelines; a regulatory affairs office might maintain skills aligned with its jurisdiction's submission requirements. Federated skill repositories, combined with versioning and provenance tracking, would allow the collection to scale while preserving institutional specificity and accountability.
 
\textbf{Context Engineering: Skills, RAG, and MCP as Complementary Layers.}
An important architectural question for Vibe Medicine is how the skill-based paradigm relates to Retrieval-Augmented Generation (RAG) and the Model Context Protocol (MCP). RAG grounds model outputs in retrieved documents and reduces hallucination over large static knowledge bases such as clinical guidelines, but it lacks reasoning and the ability to execute actions. MCP provides a standardized interface for discovering and invoking external tools, APIs, and databases at runtime, enabling transactional operations such as querying a live PACS server or writing results to an EHR, but specifies only how to connect, not what domain knowledge to apply \cite{AnthropicMCP2024}. Skills occupy a distinct layer: they encode behavioral expertise, domain workflows, analytical procedures, output conventions, and quality standards that neither retrieval nor tool invocation alone can provide. A skill such as \texttt{tooluniverse-variant-interpretation} \cite{Gaoetal2025ToolUniverse,gao2025txagent} specifies the full interpretive protocol, including evidence aggregation across multiple databases, ACMG classification logic, and structured clinical reporting, which RAG cannot do and MCP does not address. A well-designed medical skill may internally orchestrate MCP-style tool calls and RAG-style retrieval, then synthesize results using domain-specific reasoning present only in the skill specification itself. Future work should formalize this compositional relationship and develop evaluation frameworks that assess integrated skill-level output quality, not only retrieval accuracy or tool invocation success.
 
\textbf{AutoResearch.}
Another promising direction is AutoResearch, in which the agent carries out an end-to-end research workflow from a user's high-level idea. The human provides only a general research objective; the agent invokes planning skills to decompose it into a structured research agenda, then executes and continually refines that agenda across the full research lifecycle, typically including survey, ideation, experiment, publication, and promotion. At each stage, the agent calls the appropriate domain skills, such as literature analysis, data processing, hypothesis testing, result summarization, manuscript drafting, and dissemination support, functioning more like a research collaborator than a single-purpose assistant. Outputs from one stage feed back into the planning module to revise priorities and adjust downstream actions, enabling a closed-loop process that is adaptive, goal-driven, and increasingly autonomous. By lowering the technical barrier to entry, such human-AI co-work systems could extend the reach of biomedical research to independent researchers and institutions in low-resource areas.

\section{Conclusion}
\label{sec:conc}

This paper introduced Vibe Medicine, a human-AI co-work paradigm that redefines biomedical research. In this paradigm, clinicians and researchers guide skill-augmented AI agents through natural language to execute complex, multi-step biomedical research workflows. Vibe Medicine repositions the researcher as a research director who focuses on the big picture while the AI agent handles implementation details. By doing this, Vibe Medicine allows more accessible high-quality research, especially for independent researchers and institutions in low-resource areas. To support this, the domain knowledge of Vibe Medicine is embedded in the OpenClaw medical skills collection, which provides more than 1,000 curated skills spanning the full breadth of biomedical research. We described the architecture of this collection, surveyed its capabilities across ten domains, and presented case studies that demonstrate how skills can be composed into end-to-end research workflows that would traditionally require a multidisciplinary team. This paper also identified substantial challenges that should be addressed before Vibe Medicine can be responsibly adopted at scale. Hallucination, misinformation, privacy, and the risk of over-reliance (the Vibe Medicine Hangover) all demand critical evaluation and human oversight. Future directions should focus on self-improving agent systems, clinically verified skill creation, and data governance. With these in place, Vibe Medicine will enable more reliable, trustworthy, and clinically integrated agent-assisted research that advances research and technological equity and reduces health care resource disparities.

\bibliographystyle{cas-model2-names}

\bibliography{cas-refs}

\end{document}